\documentclass[conference]{IEEEtran}
\IEEEoverridecommandlockouts
\usepackage{cite}
\usepackage{amsmath,amssymb,amsfonts}
\usepackage{algorithmic}
\usepackage{graphicx}
\usepackage{textcomp}
\usepackage[utf8]{inputenc}
\usepackage{xcolor}
\def\BibTeX{{\rm B\kern-.05em{\sc i\kern-.025em b}\kern-.08em
    T\kern-.1667em\lower.7ex\hbox{E}\kern-.125emX}}

\usepackage{graphicx}
\usepackage{multirow}
\usepackage{bbding}
\usepackage{enumitem}
\usepackage[colorlinks,linkcolor=red,anchorcolor=black,citecolor=green]{hyperref}  
\usepackage{url}            
\usepackage{booktabs}       

\usepackage{tikz}
\definecolor{low}{RGB}{185, 101, 71}
\definecolor{middle low}{RGB}{248, 202, 155}
\definecolor{middle}{RGB}{211, 232, 158}
\definecolor{middle high}{RGB}{138, 191, 104}
\definecolor{high}{RGB}{92, 144, 77}

\definecolor{clear sky}{RGB}{79,253,199}
\definecolor{thick cloud}{RGB}{77,2,115}
\definecolor{thin cloud}{RGB}{251,255,41}
\definecolor{cloud shadow}{RGB}{221,53,223}
    
\begin{document}

\title{Knowledge Transfer and Domain Adaptation for Fine-Grained Remote Sensing Image Segmentation}

\author{\IEEEauthorblockN{Shun~Zhang$^{1,*}$, Xuechao Zou$^{2,*}$, Kai Li$^{3}$, Congyan Lang$^{2}$, Shiying Wang$^{1,\dagger}$, Pin Tao$^{1,3,\dagger}$, Tengfei Cao$^{1}$}
\IEEEauthorblockA{$^1$School of Computer Technology and Applications, Qinghai University, Xining, China \\
$^2$School of Computer Science and Technology, Beijing Jiaotong University, Beijing, China \\
$^3$Department of Computer Science  and Technology, Tsinghua University, Beijing, China \\
\small{cs.shunzhang@foxmail.com}
}
}

\maketitle

\begin{abstract}
Fine-grained remote sensing image segmentation is essential for accurately identifying detailed objects in remote sensing images. Recently, vision transformer models (VTMs) pre-trained on large-scale datasets have demonstrated strong zero-shot generalization. However, directly applying them to specific tasks may lead to domain shift. We introduce a novel end-to-end learning paradigm combining knowledge guidance with domain refinement to enhance performance. We present two key components: the Feature Alignment Module (FAM) and the Feature Modulation Module (FMM). FAM aligns features from a CNN-based backbone with those from the pretrained VTM's encoder using channel transformation and spatial interpolation, and transfers knowledge via KL divergence and L2 normalization constraint. FMM further adapts the knowledge to the specific domain to address domain shift. We also introduce a fine-grained grass segmentation dataset and demonstrate, through experiments on two datasets, that our method achieves a significant improvement of 2.57 mIoU on the grass dataset and 3.73 mIoU on the cloud dataset. The results highlight the potential of combining knowledge transfer and domain adaptation to overcome domain-related challenges and data limitations. The project page is available at \url{https://xavierjiezou.github.io/KTDA/}.
\end{abstract}

\begin{IEEEkeywords}
Fine-grained image segmentation, remote sensing, knowledge transfer, domain adaptation, vision transformer
\end{IEEEkeywords}

\section{Introduction}
\label{intro}

Remote sensing image segmentation is vital for applications such as environmental monitoring, agricultural surveys, and urban planning~\cite{dong2023large,han2021fine,li2021multiattention}, particularly in the precise identification of fine-grained objects like grass, clouds, and crops at various levels. As the volume and complexity of remote sensing data continue to increase, traditional segmentation methods~\cite{fcn,unet,pspnet,deeplabv3+} struggle to handle intricate and fine-grained objects. Consequently, enhancing the accuracy of remote sensing image segmentation, particularly for fine-grained tasks, has become an urgent challenge.

Recent advances in deep learning, especially convolutional neural networks (CNNs), have made significant strides in remote sensing image processing~\cite{pmaa, diffcr}. More recently, vision transformers (VTMs)~\cite{wang2024sam,dinov2,clip} have shown great potential in remote sensing image analysis due to their ability to model long-range dependencies~\cite{li2024iianet}. Large-scale pre-trained VTM models have demonstrated strong generalization across various vision tasks, offering valuable prior knowledge for remote sensing image segmentation. However, despite their success on large-scale datasets~\cite{imagenet}, these models still face challenges in performing well on domain-specific remote sensing tasks, particularly in fine-grained object semantic segmentation.

There are significant domain differences between remote sensing images and general visual images. For example, remote sensing images have unique imaging characteristics, sensor types, resolutions, and scales, and the objects in these images often exhibit textures and structures that differ from those found in natural images. These differences make it difficult to directly apply pre-trained models designed for general visual tasks to remote sensing image segmentation, often leading to performance degradation when processing remote sensing data. Additionally, fine-grained objects in these images typically have small sizes and subtle details, further complicating the segmentation task.

To address these challenges, we propose an end-to-end learning framework that integrates knowledge transfer with domain adaptation. The framework features two key components: the Feature Alignment Module (FAM) and the Feature Modulation Module (FMM). FAM aligns features extracted by a CNN backbone with those from a pre-trained VTM encoder through channel transformation and spatial interpolation, enabling knowledge transfer via KL divergence, L2 normalization and auxiliary constraints. FMM then further adapts the transferred knowledge to the target domain to mitigate domain shift. Experimental results demonstrated that our method significantly improves segmentation performance, underscoring the effectiveness of combining knowledge transfer and domain adaptation to address domain shift and data scarcity. In summary, our contributions are as follows:

\begin{itemize}
    \item We present an end-to-end learning paradigm that integrates knowledge transfer and domain adaptation for fine-grained remote sensing image segmentation. This framework effectively transfers knowledge from pre-trained models and adapts it to specific target domains.
    \item We develop two key components: the Feature Alignment Module (FAM), and the Feature Modulation Module (FMM). FAM aligns features extracted by a CNN backbone with those from a pre-trained VTM encoder, ensuring a stable transition of feature distributions. FMM then further adapts features to mitigate domain shift.
    \item We introduce the first fine-grained grass dataset for remote sensing image segmentation that addresses challenges such as boundary ambiguity and misclassification.
\end{itemize}

\section{Related Work}

\subsection{Fine-Grained Remote Sensing Image Segmentation}

Fine-grained segmentation refines semantic segmentation by distinguishing subcategories within classes, such as grass, shrubs, and trees under vegetation, which is critical for tasks like ecological monitoring. CNN-based models~\cite{fcn, deeplabv3+, pspnet} leverage multi-scale for local modeling, while transformer-based models~\cite{segformer, mask2former, dinov2} enhance long-range dependency modeling. For tasks like cloud segmentation, methods such as~\cite{cdnetv1, cdnetv2} address subtle distinctions. However, challenges persist due to limited fine-grained vegetation datasets, especially for grasslands, and the reliance on coarse classifications, highlighting the need for more advanced techniques.

\subsection{Knowledge Distillation and Transfer Learning}

Knowledge distillation, first introduced in~\cite{hinton2015distilling}, is a form of transfer learning where a smaller (student) model mimics a larger (teacher) model’s behavior using soft targets — probability distributions from the teacher. Recent works like MobileSAM \cite{mobileSam} and EfficientSAM \cite{xiong2024efficientsam} apply this technique to transfer knowledge from SAM’s image encoder~\cite{sam} to lightweight models, achieving high performance and fast inference for real-time applications. This aids in model compression and enhances generalization by transferring knowledge from a larger model. Transfer learning, a broader technique, involves adapting a pre-trained model to improve performance on related tasks, particularly with limited labeled data. This includes domain adaptation, where models are fine-tuned for specialized tasks, such as remote sensing image segmentation, by transferring knowledge from one domain to another.

\subsection{Vision Transformers and Domain Adaptation}
Vision Transformers (ViTs)~\cite{vit}, including SAM~\cite{sam}, DINOv2~\cite{dinov2}, and MAE~\cite{mae}, demonstrate strong generalization across visual tasks due to their attention-based architectures, which capture long-range dependencies and learn transferable features. However, when applied to specialized tasks, such as remote sensing image segmentation (e.g., grass, cloud detection), their performance may suffer due to the mismatch between general pre-training datasets and domain-specific features. Remote sensing images often contain unique visual attributes (e.g., lighting variations, complex spatial structures) not present in conventional datasets. Domain adaptation~\cite{cloud-adapter} addresses this by fine-tuning ViTs for specialized tasks, transferring knowledge from a source domain (e.g., general datasets) to a target domain (e.g., remote sensing images).

\section{Method}

\begin{figure*}[t]
    \centering
    \includegraphics[width=0.85\linewidth]{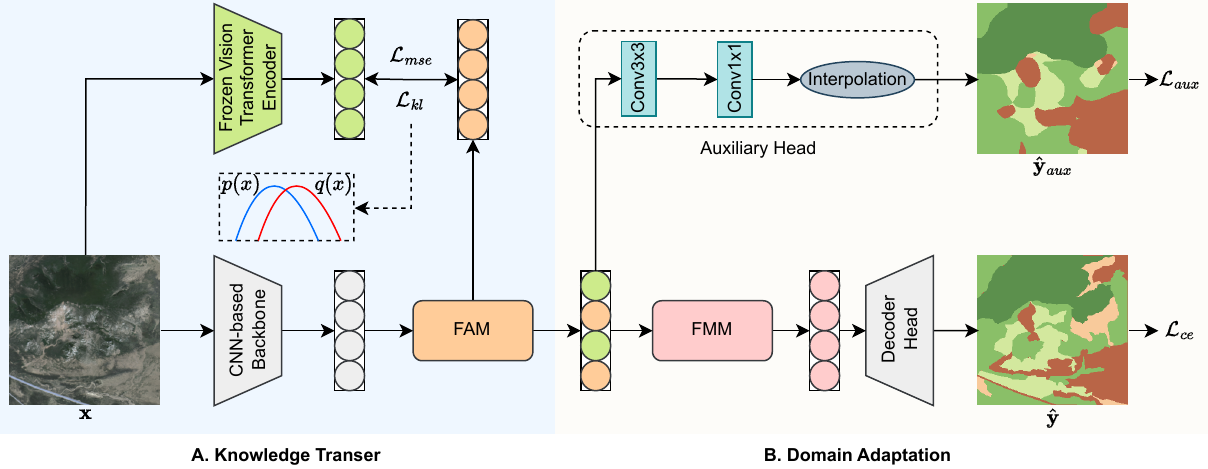}
    \caption{Overview of the proposed framework that integrates knowledge transfer and domain adaptation for fine-grained remote sensing image segmentation.}
    \label{fig: framework}
\end{figure*}

We propose an end-to-end learning paradigm that integrates knowledge transfer with domain adaptation. As illustrated in Fig.~\ref{fig: framework}, the framework consists of two parts: knowledge transfer and domain adaptation. First, knowledge is transferred from a vision transformer model pre-trained on a general domain to a CNN-based backbone. Next, domain adaptation is employed to mitigate domain shifts, thereby enhancing fine-grained segmentation performance. Overall, our approach effectively leverages the benefits of knowledge guidance and domain refinement.

\begin{figure}
    \centering
    \includegraphics[width=0.90\linewidth]{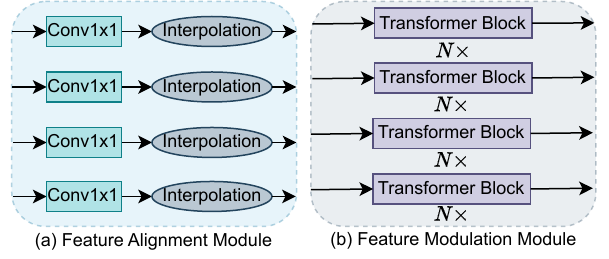}
     \caption{Detailed structure of the FAM and FMM.}
    \label{fig: module}
\end{figure}

\subsection{Knowledge Transfer} \label{sec:pipeline}
Knowledge transfer enables the CNN-based backbone to effectively learn from the encoder of the frozen vision transformer model (VTM), as shown in Fig.~\ref{fig: framework}A. Specifically, Knowledge transfer is achieved through the combination of feature alignment and loss calculation. In the Feature Alignment Module (FAM), the multi-scale features from the backbone are aligned with those of the VTM by adjusting their channel dimensions and resizing the feature maps to match the VTM’s spatial resolution.

\subsubsection{Feature Alignment Module} 
The FAM is a simple yet highly effective component. Given an input image \( \mathbf{{x}} \in \mathbb{R}^{C \times H \times W} \), where \( C \) represents the number of channels, \( H \) is the height, and \( W \) is the width, the backbone generates multi-scale features: $
\{{\mathbf{F}_1}, {\mathbf{F}_2}, \dots, {\mathbf{F}_n}\}, {\mathbf{F}_i} \in \mathbb{R}^{C_i \times H_i \times W_i}$, $n$ denotes the number of features. Here, we set $n=4$ as the default configuration. FAM begins by adjusting the channel of each feature using convolutions with a kernel size of \( 1\times 1\):

\begin{equation}
\mathbf{F}'_i = \text{Conv}_{i}(\mathbf{F}_i),
\end{equation}
where \( \text{Conv}_{i} \) represent the $i$-th convolution operation, $\mathbf{F}'_i\in\mathbb{R}^{C'_i \times H_i \times W_i}$ denotes the features after channel transformation. Next, an interpolation algorithm is applied to resize the feature maps, aligning their spatial resolution with that of the VTM's:

\begin{equation}
\mathbf{F}''_i = \varnothing(\mathbf{F}'_i, H_{\text{vtm}}, W_{\text{vtm}}),
\end{equation}
where \(\mathbf{F}''_i\in\mathbb{R}^{C'_i \times H_\text{vtm} \times W_\text{vtm}} \) denotes the \(i\)-th feature map, and \( \varnothing \) denotes bilinear interpolation. This straightforward alignment process facilitates effective transfer learning by enabling knowledge transfer from the VTM to the backbone.

\subsubsection{Loss Function} 
The loss function for knowledge transfer is designed to minimize the discrepancy between the feature representations of the backbone and the VTM after feature alignment. The knowledge transfer loss is computed by combining the mean squared error \( \mathcal{L}_{\text{mse}} \) and kullback-leibler divergence \( \mathcal{L}_{\text{kl}} \) losses to enhance model optimization and regularization in a stable transition. The loss at each feature scale is calculated using both \( \mathcal{L}_{\text{mse}} \) and \( \mathcal{L}_{\text{kl}} \) as follows:

\begin{equation}
\mathcal{L}_{\text{mse}} = \frac{1}{n} \sum_{i=1}^{n} \left\| \mathbf{F}''_i - \mathbf{F}_i^{\text{vtm}} \right\|_2^2,
\end{equation}
\begin{equation}
\mathcal{L}_{\text{kl}} = \frac{1}{n} \sum_{i=1}^{n} \text{KL}(\mathbf{F}''_i \| \mathbf{F}_i^{\text{vtm}}),
\end{equation}
where \(\mathbf{F}_i^\text{VTM} \in \mathbb{R}^{C'_i \times H_\text{vtm} \times W_\text{vtm}}\) represents the feature map from the VTM. The terms \(\mathcal{L}_{\text{mse}}\) and \(\mathcal{L}_{\text{kl}}\) are computed by averaging over all scales to ensure consistent alignment between the backbone and VTM across different resolutions. The combined loss, \( \mathcal{L}_{\text{kt}} \), is a weighted sum of \( \mathcal{L}_{\text{mse}} \) and \( \mathcal{L}_{\text{kl}} \):

\begin{equation}
\mathcal{L}_{\text{kt}} = \lambda_{\text{mse}} \cdot \mathcal{L}_{\text{mse}} + \lambda_{\text{kl}} \cdot \mathcal{L}_{\text{kl}},
\end{equation}
where \( \mathcal{L}_{\text{kt}} \) denotes the knowledge transfer loss, and we use \( \lambda_{\text{mse}} = \lambda_{\text{kl}} = 0.5 \) as default weights.

\subsection{Domain Adaptation}

Domain adaptation (see~Fig.~\ref{fig: framework}B) involves a feature modulation module built upon transformer blocks to modulate the feature distribution from the source (VTM) to target (segmentation) domains. The module processes the input features \( \mathbf{F}''_i \) through \(N\) sequential transformer blocks, progressively adapting the feature distribution to better match the target domain. Then, a dual-decoder head architecture is employed to improve performance, with the primary head and auxiliary head optimized through a weighted combination of losses.

\subsubsection{Feature Modulation Module}
The FMM is also a simple yet effective approach for adapting the feature distribution from a general domain to a target domain using transformer blocks. Given the feature map \( \mathbf{F}''_i \) from the FAM, the module applies \(N\) transformer blocks to modulate the features:

\begin{equation}
\mathbf{F}_j = \text{TB}_N(\mathbf{F}''_i),
\end{equation}
where $\text{TB}$ denotes the transformer block,  each $\text{TB}$ consists of a self-attention mechanism and a feedforward neural network. The final output is \( \mathbf{F}_j \in \mathbb{R}^{C'_i \times H_\text{vtm} \times W_\text{vtm}} \), $j \in \{1, \dots, n\}$.

\subsubsection{Loss Function} 
The loss function for domain adaptation in semantic segmentation is designed based on the outputs from the FAM and FMM. After passing through \(N\) transformer blocks, the feature map \( \mathbf{F}_j\) is processed by the decoder head to produce the predicted segmentation output \(\hat{\mathbf{y}}\in \mathbb{R}^{1 \times H \times W}\). In parallel, $\mathbf{F}''_i$ feeds the auxiliary head to generate \(\hat{\mathbf{y}}_{\text{aux}}\in \mathbb{R}^{1 \times H \times W}\). The auxiliary head consists of two convolution operations and an interpolation.

The domain adaptation loss, \( \mathcal{L}_{da} \), consists of two parts: the cross-entropy loss \( \mathcal{L}_{\text{ce}} \) calculated from the decoder head's output, and the auxiliary loss \( \mathcal{L}_{\text{aux}} \) calculated from the auxiliary head. These components are defined as follows:

\begin{equation}
\mathcal{L}_{ce} = \text{CE}(\hat{\mathbf{y}}, \mathbf{y}),
\quad
\mathcal{L}_{aux} = \text{CE}(\hat{\mathbf{y}}_{aux}, \mathbf{y}),
\end{equation}
where \(\mathbf{y}\in \mathbb{R}^{1 \times H \times W} \) is the ground truth. \( \mathcal{L}_{da} \) is as follows:

\begin{equation}
\mathcal{L}_{da} = \lambda_{\text{ce}} \cdot \mathcal{L}_{ce} + \lambda_{\text{aux}} \cdot \mathcal{L}_{aux},
\end{equation}
where, we use \( \lambda_{\text{ce}} = 1.0 \) and \( \lambda_{\text{aux}} = 0.4 \) as default weights.

The total loss for our end-to-end learning paradigm combines both the knowledge transfer and the domain adaptation:

\begin{equation}
\mathcal{L}_{total} = \lambda_{\text{kt}} \cdot \mathcal{L}_{kt} + \lambda_{\text{da}} \cdot \mathcal{L}_{da},
\end{equation}
where \( \lambda_{\text{kt}} = \lambda_{\text{da}} =1.0 \). The total loss function allows the model to optimize knowledge transfer and domain adaptation simultaneously, improving segmentation performance.

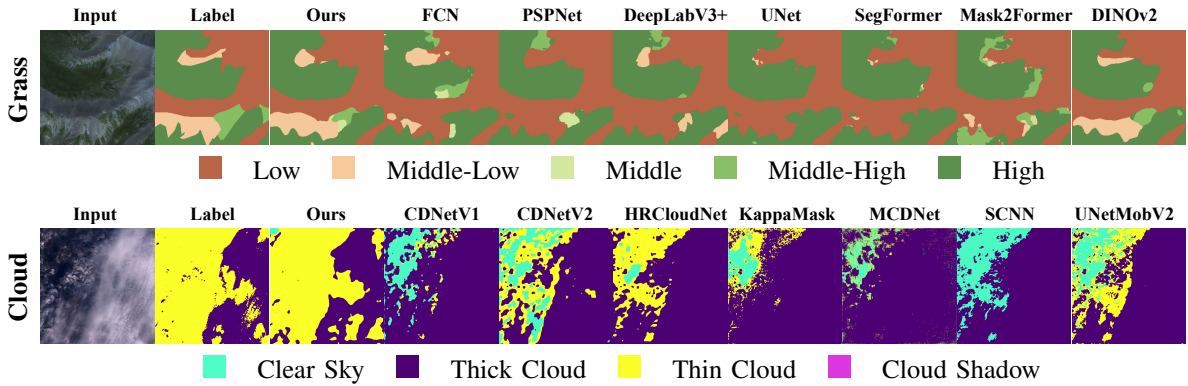
\begin{figure*}[]
\centering
\includegraphics[width=0.85\linewidth]{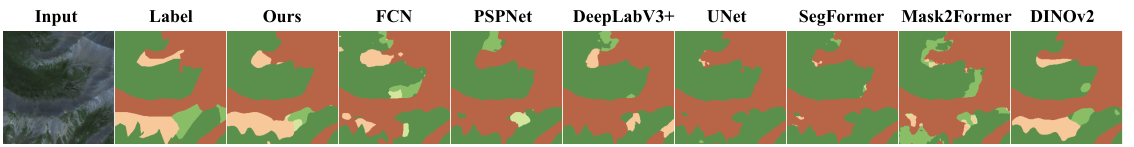}
\begin{tikzpicture}[overlay]
    \node[anchor=west, xshift=1.0cm,rotate=90] at (-16.8, 0.2) {\textbf{Grass}};
\end{tikzpicture}
\begin{tikzpicture}
    \node {
        \begin{tabular}{cccccccccc}
            \tikz\fill[low] (0,0) rectangle (0.3,0.3); & Low &
            \tikz\fill[middle low] (0,0) rectangle (0.3,0.3); & Middle-Low &
            \tikz\fill[middle] (0,0) rectangle (0.3,0.3); & Middle &
            \tikz\fill[middle high] (0,0) rectangle (0.3,0.3); & Middle-High &
            \tikz\fill[high] (0,0) rectangle (0.3,0.3); & High \\
        \end{tabular}
    };
\end{tikzpicture}
\includegraphics[width=0.85\linewidth]{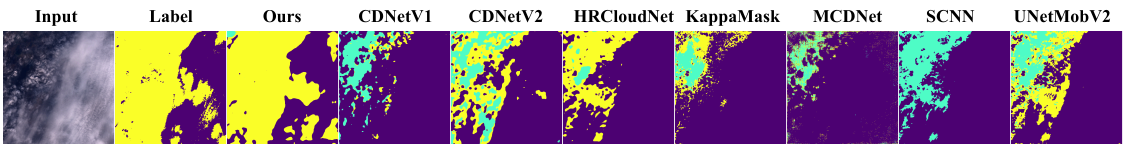}
\begin{tikzpicture}[overlay]
    \node[anchor=west, xshift=1.0cm,rotate=90] at (-16.8, 0.2) {\textbf{Cloud}};
\end{tikzpicture}
\begin{tikzpicture}
    \node {
        \begin{tabular}{cccccccc}
            \tikz\fill[clear sky] (0,0) rectangle (0.3,0.3); & Clear Sky &
            \tikz\fill[thick cloud] (0,0) rectangle (0.3,0.3); & Thick Cloud &
            \tikz\fill[thin cloud] (0,0) rectangle (0.3,0.3); & Thin Cloud &
            \tikz\fill[cloud shadow] (0,0) rectangle (0.3,0.3); & Cloud Shadow \\
        \end{tabular}
    };
\end{tikzpicture}
\caption{Comparison of visualization segmentation results of different models on the fine-grained grass and cloud segmentation datasets.}
\label{fig: grass}
\end{figure*}

\section{Experiment}

\subsection{Datasets}

\subsubsection{Fine-Grained Grass Segmentation}

We have developed a fine-grained grass dataset to improve the accuracy of grassland extraction in complex terrains. Existing datasets~\cite{helber2019eurosat, GID2020, Boguszewski_2021_CVPR, kemker2018algorithms} lack detailed labeling, limiting their ability to address challenges like boundary ambiguity and misclassification. Our dataset fills this gap by offering a more refined classification, which enhances grass segmentation and contributes to a better understanding of grass ecosystems.

The dataset was created using high-resolution (8m) satellite imagery from the Gaofen series (Gaofen-2 and Gaofen-6), captured in 2019 over Maduo County, China, located in the Yellow River source area. This region is known for its high-altitude, alpine grasslands, and complex terrain, with coordinates between 33°50'–35°40' N latitude and 96°50'–99°20' E longitude. We collected two 13,872 × 13,150-pixel images from Gaofen-6 and two 7,300 × 6,905-pixel images from Gaofen-2, all containing red, green, and blue spectral bands. These images provide critical information for fine-grained grass extraction. Labeling was assisted by the X-AnyLabeling~\footnote{\url{https://github.com/CVHub520/X-AnyLabeling}} tool, supplemented with manual refinements to ensure high accuracy. Grassland coverage was classified into five levels, based on national grassland survey standards: low coverage (\textless10\%), medium-low coverage (10\%-25\%), medium coverage (25\%-50\%), medium-high coverage (50\%-75\%), and high coverage (\textgreater75\%), as shown in Tab.~\ref{tab: dataset}. The final dataset comprises 1,151 pairs of 256$\times$256 patches, split into training and testing sets with an 8:2 ratio. This dataset, with its detailed and accurate labeling, is a valuable resource for advancing remote sensing applications, particularly in ecologically sensitive and high-altitude regions like the Yellow River source area.

\begin{table}[h!]
\centering
\caption{Statistics of grass coverage levels and pixel count.}
\setlength{\tabcolsep}{12pt}
\begin{tabular}{cccc}
\toprule
\textbf{Level} & \textbf{Coverage}        & \textbf{Range (\%)} & \textbf{Pixels} \\ \midrule
I              & Low         & $<10$            & 12,680,802      \\
II             & Middle-Low           & 10-25         & 13,598,315   \\
III            & Middle               & 25-50         & 9,744,128       \\
IV             & Middle-High          & 50-75        & 7,919,316       \\
V              & High        & $> 75$           & 31,489,375      \\ 
\bottomrule
\label{tab: dataset}
\end{tabular}
\end{table}


\subsubsection{Fine-Grained Cloud Segmentation}

The dataset\cite{cloud2017datasets} consists of 96 terrain-corrected (Level-1T) scenes from Landsat 8 OLI and TIRS, covering diverse biomes. This variety supports cloud detection and removal in complex environments. The dataset includes manually generated cloud masks with pixel-level annotations for cloud shadow, clear sky, thin clouds, and cloud areas. Each scene is cropped into 512×512 pixel patches and split into training, validation, and test sets (6:2:2 ratio). It is a valuable resource for training and evaluating fine-grained cloud segmentation models across various terrains.



\subsection{Implementation Details}
\subsubsection{Hyperparameter and Training Settings}
All experiments were conducted using the MMSegmentation~\footnote{\url{https://github.com/open-mmlab/mmsegmentation}}, running on 8 NVIDIA A100 GPUs. Our experiment employed UPerNet~\cite{upernet} as the default decoder head and FCN~\cite{fcn} as the auxiliary head, unless specified otherwise. The experiments were trained for a maximum of 23,000 iterations. The optimizer used for training was AdamW~\cite{adamw} with an initial learning rate of \( 1 \times 10^{-3} \), with a custom learning rate schedule. The learning rate started with a linear warm-up over the first 1,150 iterations, followed by a PolyLR scheduler with a minimum eta value of 0.0 and a power parameter set to 0.9. The learning rate decay continued until the end of the iteration. The batch size was set to 4.

\subsubsection{Baselines and Evaluation Metrics}

To comprehensively evaluate the performance of our proposed method, we compare it with several widely recognized segmentation models. For the grass datasets, we use the following models as baselines: FCN~\cite{fcn}, PSPNet~\cite{pspnet}, DeepLabV3+~\cite{deeplabv3+}, SegFormer~\cite{segformer}, Mask2Former~\cite{mask2former}, DINOv2~\cite{dinov2}, and U-Net~\cite{unet}. For the cloud dataset, we use models specifically designed for cloud segmentation: SCNN~\cite{scnn}, CDNetV1~\cite{cdnetv1}, CDNetV2~\cite{cdnetv2}, MCDNet~\cite{mcdnet}, UNetMobV2~\cite{cloudsen12_high}, HRCloudNet~\cite{hrcloudnet}, and KappaMask~\cite{kappamask}. All of feature extraction networks are pretrained on the ImageNet~\cite{imagenet}, ensuring a fair comparison. We evaluate segmentation performance using three standard metrics: Mean Intersection over Union (mIoU), F1 Score, and Overall Accuracy (OA), all reported as percentages (\%).


\subsection{Ablation and Discussion}

\subsubsection{Study of Loss on Knowledge Transfer}

Tab.~\ref{tab: loss_ablation} shows the impact of every loss functions on knowledge transfer from the Frozen Vision Transformer Encoder~\cite{dinov2}. The best performance is achieved by combining all loss items, reaching a mIoU of 50.81\%, OA of 74.10\%, and F1 score of 64.91\%. Among these components,~\( \mathcal{L}_{\text{kl}} \) demonstrates the most significant impact, as evidenced by the substantial performance drop when it is removed (mIoU decreases by 7.13\%). Removing the \( \mathcal{L}_{\text{aux}} \) results in a relatively minor performance decline (mIoU drops by 1.21\%), while excluding \( \mathcal{L}_{\text{mse}} \) leads to a moderate decrease (mIoU reduces by 1.63\%). These results indicate that \( \mathcal{L}_{\text{kl}} \) plays a crucial role in effective knowledge transfer from the knowledge of the pre-trained vision transformer encoder.

\begin{table}[]
\centering
\setlength{\tabcolsep}{2mm}
\caption{Study on the presence of each loss function of our method.}
\label{tab: loss_ablation}
\begin{tabular}{cc|cc|ccc}
\toprule
\multicolumn{2}{c|}{Knowledge Transfer} & \multicolumn{2}{c|}{Domain Adaptation} & \multicolumn{3}{c}{Metrics} \\
\cmidrule(lr){1-2} \cmidrule(lr){3-4} \cmidrule(lr){5-7}
\(\mathcal{L}_{\text{kl}}\) & \(\mathcal{L}_{\text{mse}}\) & \(\mathcal{L}_{\text{aux}}\) & \(\mathcal{L}_{\text{ce}}\) & mIoU ↑ & OA ↑  & F1 ↑  \\
\midrule
\(\checkmark\)  & \(\checkmark\)   & \(\checkmark\)    & \(\checkmark\)  & \textbf{50.81}  & \textbf{74.10} & \textbf{64.91} \\
\(\checkmark\)  & \(\checkmark\)   & \(\times\)    & \(\checkmark\)  & 49.60  & 73.30 & 63.75 \\
\(\checkmark\)  & \(\times\)   & \(\checkmark\)    & \(\checkmark\)  & 49.18  & 72.78 & 63.34 \\
\(\times\)  & \(\checkmark\)   & \(\checkmark\)    & \(\checkmark\)  & 43.68  & 68.71 & 57.41 \\ 
\bottomrule
\end{tabular}
\end{table}

\subsubsection{Impact of FAM on Knowledge Transfer}

\begin{table}[]
\centering
\setlength{\tabcolsep}{5.5mm}
\caption{Effect of FAM on knowledge transfer.}
\label{tab: loss_fam}
\begin{tabular}{l|ccc}
\toprule
FAM            & mIoU ↑         & OA ↑           & F1 ↑           \\ \midrule
Single-Scale ($n=1$) & 47.88          & 71.43          & 62.25          \\
Multi-Scale ($n=4$)  & 49.00             & 73.27          & 62.89          \\ \midrule
\(\times\)              & 47.57          & 71.54          & 61.7           \\
\(\checkmark\)~(w/o loss)      & 48.64          & 73.10           & 62.51          \\
\(\checkmark\)~(w/ loss)      & \textbf{49.29} & \textbf{73.19} & \textbf{63.38} \\ \bottomrule
\end{tabular}
\end{table}

Tab.~\ref{tab: loss_fam} illustrates the effect of the Feature Alignment Module and the knowledge transfer loss, which combines  \( \mathcal{L}_{\text{kl}} \),  \( \mathcal{L}_{\text{mse}} \), and   \( \mathcal{L}_{\text{aux}} \), on knowledge transfer. Incorporating FAM with multi-scale outputs consistently improves performance, with mIoU increasing from 47.88\% to 49.00\%. When FAM is combined with the  \( \mathcal{L}_{\text{aux}} \), the best performance is achieved, with mIoU reaching 49.29\%, OA of 73.19\%, and F1 score of 63.38\%. These results highlight that the improvement is not solely due to an increase in the number of parameters introduced by FAM, but rather the synergistic effect of the knowledge transfer loss, which plays a crucial role in enhancing knowledge transfer and improving segmentation accuracy.

\subsubsection{Impact of FMM on Domain Adaptation}

\begin{table}[]
\centering
\setlength{\tabcolsep}{8mm}
\caption{Study on the number of transformer blocks in the FMM.}
\label{tab: fmm}
\begin{tabular}{c|ccc}
\toprule
$N$ & mIoU ↑         & OA ↑           & F1 ↑           \\ \midrule
0   & 48.82          & 72.76          & 62.95          \\ \midrule
1   & 49.90          & 73.21          & 64.08          \\
2   & 50.57          & 73.62          & 64.78          \\
3   & 49.64          & 73.27          & 63.78          \\
4   & \textbf{50.86} & \textbf{74.26} & \textbf{65.01} \\ \bottomrule
\end{tabular}
\end{table}

Tab.~\ref{tab: fmm} shows the effect of the Feature Modulation Module on domain adaptation. As the number of Transformer Blocks in FMM increases, with the best results achieved at $N$=4, reaching mIoU of 50.86\%. The performance slightly declines beyond $N$=2, indicating diminishing returns. These results highlight the effectiveness of FMM in enhancing domain adaptation, with $N$=4 providing the optimal performance.

\subsection{Comparison with State-of-the-Art Methods}

\subsubsection{Quantitative Analysis}

\begin{table}[]
\centering
\setlength{\tabcolsep}{5.5mm}
\caption{Performance comparison of different methods on our fine-grained grass segmentation dataset.}
\label{tab: main_table}
\begin{tabular}{l|ccc}
\toprule
Method      & mIoU~$\uparrow$         & OA~$\uparrow$           & F1~$\uparrow$           \\ \midrule
FCN~\cite{fcn}         & 47.47          & 67.85          & 61.99          \\
PSPNet~\cite{pspnet}      & 47.95          & 69.12          & 62.55          \\
DeepLabV3+~\cite{deeplabv3+}  & 47.95          & 68.97          & 62.50          \\
UNet~\cite{unet}        & 48.17          & 69.77          & 62.34          \\
SegFormer~\cite{segformer}   & 48.29          & 68.93          & 62.82          \\
Mask2Former~\cite{mask2former} & 44.93          & 65.90          & 58.91          \\
DINOv2~\cite{dinov2}      & 47.57          & 71.54          & 61.70          \\ \midrule
Ours        & \textbf{50.86} & \textbf{74.26} & \textbf{65.01} \\ \bottomrule
\end{tabular}
\end{table}

\begin{table}[]
\centering
\setlength{\tabcolsep}{5.5mm}
\caption{Performance comparison of different methods on the fine-grained cloud segmentation dataset.}
\label{tab: main_l8}
\begin{tabular}{l|ccc}
\toprule
Method     & mIoU~$\uparrow$  & OA~$\uparrow$    & F1~$\uparrow$    \\ \midrule
MCDNet~\cite{mcdnet}     & 33.85 & 69.75 & 42.76 \\
SCNN~\cite{scnn}       & 32.38 & 71.22 & 52.41 \\
CDNetv1~\cite{cdnetv1}    & 34.58 & 68.16 & 45.80  \\
KappaMask~\cite{kappamask}  & 42.12 & 76.63 & 68.47 \\
UNetMobv2~\cite{cloudsen12_high}  & 47.76 & 82.00  & 56.91 \\
CDNetv2~\cite{cdnetv2}    & 43.63 & 78.56 & 70.33 \\
HRCloudNet~\cite{hrcloudnet} & 43.51 & 77.04 & 71.36 \\ \midrule
Ours       & \textbf{51.49} & \textbf{83.55} & \textbf{60.08} \\ \bottomrule
\end{tabular}
\end{table}

We perform a comprehensive evaluation of our method by comparing it with state-of-the-art approaches on two fine-grained segmentation tasks: grass segmentation and cloud segmentation. The evaluation metrics include mean Intersection over Union (mIoU), overall accuracy (OA), and F1-score. The detailed performance comparisons are shown in Tab.~\ref{tab: main_table} and Tab.~\ref{tab: main_l8}.

For grass segmentation, As shown in Tab.~\ref{tab: main_table}, our method achieves the best performance in all metrics (mIoU: 50.86, OA: 74.26, F1: 65.01), outperforming SegFormer~\cite{segformer}, the next best method, by 2.57\% in mIoU. Transformer-based models like Mask2Former~\cite{mask2former} and DINOv2~\cite{dinov2} underperform, likely due to their reliance on large-scale datasets for effective training, which is constrained by the limited size of our dataset. In contrast, convolutional models such as SegFormer~\cite{segformer} and UNet~\cite{unet} demonstrate better robustness in low-data regimes, highlighting the importance of dataset scale in transformer-based architectures.

Similarly, in the cloud segmentation task, our approach achieves superior results, as shown in Tab.~\ref{tab: main_l8}. We attain a mIoU of 51.49, outperforming the previous best method, HRCloudNet~\cite{hrcloudnet}, which achieves a mIoU of 43.51. While methods like UNetMobv2~\cite{cloudsen12_high} and KappaMask~\cite{kappamask} also show competitive results, they fall short in comparison to our method, with mIoU values of 47.76 and 42.12, respectively. 

Our approach thus demonstrates strong performance and significant improvements over existing methods on both the grass and cloud segmentation tasks, highlighting its effectiveness in handling fine-grained segmentation challenges.

\subsubsection{Visualization Results}

We present a qualitative comparison of our method with state-of-the-art approaches on two fine-grained segmentation tasks: grass segmentation and cloud segmentation. The visualization results are shown in Fig.~\ref{fig: grass}, where we provide segmentation outputs for various models, highlighting the differences in performance. Overall, the visualization results demonstrate the superiority of our method in producing fine-grained and accurate boundaries for both grass and cloud segmentation tasks, highlighting its effectiveness in handling data with complex and varied scenes.

\section{Conclusion}

This paper presents an innovative approach for fine-grained remote sensing image segmentation by combining knowledge transfer and domain adaptation. Through the integration of the feature alignment module and feature modulation module, our method effectively leverages pretrained vision transformers to enhance segmentation accuracy. Experiments on two datasets, including the novel fine-grained grass segmentation dataset, demonstrate significant improvements in performance, showcasing the effectiveness of knowledge transfer and domain adaptation in addressing domain shift and data limitations.



\bibliographystyle{IEEEbib}
\bibliography{main}

\end{document}